\documentclass{article}
\pdfoutput=1
\usepackage[hyphens]{url}
\usepackage{hyperref}
\usepackage{xurl}
\hypersetup{breaklinks=true,colorlinks,allcolors=blue}
\usepackage[backend=biber]{biblatex}

\usepackage{amsmath,amssymb,amsfonts}
\usepackage{algorithmic}
\usepackage{graphicx}
\usepackage{textcomp}
\usepackage[table,xcdraw]{xcolor}
\usepackage{mathtools}  
\usepackage{orcidlink}  
\usepackage{multicol}
\usepackage{authblk}
\usepackage{float}

\usepackage{geometry}
\def\BibTeX{{\rm B\kern-.05em{\sc i\kern-.025em b}\kern-.08em
    T\kern-.1667em\lower.7ex\hbox{E}\kern-.125emX}}

\begin{document}

\title{Confronting the Reproducibility Crisis:\\ A Case Study of Challenges in Cybersecurity AI}

\author{ Richard H. Moulton\,\orcidlink{0009-0005-5568-1012}\thanks{rich.moulton@trojans.dsu.edu}}
\author{ Gary A. McCully\,\orcidlink{0009-0005-6163-7720}\thanks{gary.mccully@trojans.dsu.edu}}
\author{ John D. Hastings\,\orcidlink{0000-0003-0871-3622}\thanks{john.hastings@dsu.edu}}
\affil[]{The Beacom College of Computer \& Cyber Sciences\\
  Dakota State University\\
  Madison, SD, USA}

\date{}
\maketitle

\begin{multicols}{2}
\begin{abstract}
In the rapidly evolving field of cybersecurity, ensuring the reproducibility of AI-driven research is critical to maintaining the reliability and integrity of security systems. This paper addresses the reproducibility crisis within the domain of adversarial robustness—a key area in AI-based cybersecurity that focuses on defending deep neural networks against malicious perturbations. Through a detailed case study, we attempt to validate results from prior work on certified robustness using the VeriGauge toolkit, revealing significant challenges due to software and hardware incompatibilities, version conflicts, and obsolescence. Our findings underscore the urgent need for standardized methodologies, containerization, and comprehensive documentation to ensure the reproducibility of AI models deployed in critical cybersecurity applications. By tackling these reproducibility challenges, we aim to contribute to the broader discourse on securing AI systems against advanced persistent threats, enhancing network and IoT security, and protecting critical infrastructure. This work advocates for a concerted effort within the research community to prioritize reproducibility, thereby strengthening the foundation upon which future cybersecurity advancements are built.
\end{abstract}


\section{Introduction}\label{intro}

As artificial intelligence (AI) and machine learning (ML) revolutionize the field of cybersecurity, the reliability and integrity of such technologies are more critical than ever. Although AI-driven systems are being deployed to protect critical infrastructure, analyze vast amounts of network traffic, and detect advanced persistent threats (APTs), an insidious challenge threatens to undermine these defenses: a reproducibility crisis\footnote{This is alternatively known as the replication crisis or replicability crisis.} in cybersecurity research \cite{baker20161,gundersen2020reproducibility,ioannidis2005most,open2015estimating} in which many studies' results cannot be reliably reproduced or replicated by other researchers.

The ability to reproduce and validate scientific findings is a fundamental principle of the scientific method, enabling the cumulative advancement of knowledge and the establishment of robust theories \cite{collins1992changing,goodman2016does,popper1959logic}. However, the rapid evolution of software tools, libraries, and dependencies, coupled with the inherent complexity of these methodologies, has created a significant barrier to reproducing and validating published research results \cite{donoho2010invitation,mukherjee2021fixing,peng2011reproducible,stodden2013toward}. This reproducibility crisis not only erodes scientific integrity but also hinders the practical deployment of robust models in real-world applications \cite{peng2011reproducible}. Without the ability to reliably reproduce and scrutinize research findings, it becomes challenging to build upon existing work, identify potential flaws or limitations, and foster trust in the proposed solutions \cite{sandve2013ten}.

Adversarial robustness \cite{RN62}  is the study of ensuring that deep neural networks (DNNs) \cite{lecun2015deep} maintain their functionality in the face of intentional or unintentional input perturbations. As powerful models are increasingly deployed in safety-critical applications, such as autonomous vehicles \cite{bojarski2016end,chen2015deepdriving,grigorescu2020survey}, medical diagnostics \cite{bejnordi2017diagnostic,esteva2017dermatologist,gulshan2016development,hannun2019cardiologist,rajpurkar2017chexnet}, and cybersecurity systems \cite{dahl2013large,saxe2017expose,vinayakumar2017applying}, the consequences of adversarial vulnerabilities can be severe, potentially leading to catastrophic failures and loss of life.

Within the field of cybersecurity, adversarial robustness is essential for developing resilient cyber defenses. However, the rapid evolution of AI technologies, coupled with the growing complexity of cybersecurity landscapes, has produced a challenging situation, making the reproduction and validation of research results increasingly difficult.

Consider the possibility: an adversarial robustness technique that cannot be independently verified might provide a false sense of security about an intrusion detection model. The model in question actually produces false negatives, potentially compromising critical infrastructure protection. The stakes are too high for such uncertainty in our AI-driven cyber defenses.

This paper presents a case study that highlights the challenges of reproducing state-of-the-art research in adversarial robustness. By attempting to validate the results of a comprehensive survey on certified robustness, we unveil a variety of obstacles that impede the reproduction of critical findings.

Our results serve as a microcosm of a broader issue facing the cybersecurity community: How can we trust and deploy AI systems for threat intelligence, malware analysis, or network security if we cannot reliably reproduce the research underpinning these technologies? This question is not merely academic; it has profound implications for the security of our digital infrastructure and the effectiveness of our cyber defenses.

This paper aims to spark a crucial dialogue within the cybersecurity community about the need for robust, verifiable research. We explore potential solutions, from standardized benchmarks and comprehensive documentation practices to collaborative initiatives that can enhance the reproducibility of AI research in cybersecurity.

\section{Background}

The integration of AI and ML into cybersecurity is revolutionizing threat detection, prevention, and response~\cite{yaseen2023ai}. These technologies offer unprecedented capabilities in quickly analyzing massive amounts of data, identifying complex patterns, and adapting to evolving threats~\cite{ansari2022impact,KAUR2023101804}. However, the effectiveness of AI-driven cybersecurity solutions hinges on their robustness and reliability, particularly in the face of adversarial attacks.

\subsection{Adversarial Robustness in Cybersecurity}

Adversarial robustness refers to the ability of ML models to maintain their performance and integrity when faced with maliciously crafted inputs. In the context of cybersecurity, this concept takes on paramount importance. For instance:
\begin{enumerate}
  \addtolength{\itemindent}{-.3cm}
\item Network Intrusion Detection: AI-based intrusion detection systems (IDS) must be robust against adversarial examples that could mask malicious network traffic as benign, potentially allowing APTs to go undetected \cite{yuan2024simple}.

\item Malware Analysis: Machine learning models used for malware classification need to resist adversarial perturbations that could cause malicious software to be misclassified as harmless~\cite{zhang2024robust}.

\item Phishing Detection: AI systems designed to identify phishing attempts must be resilient against subtle manipulations that could cause fraudulent emails to bypass detection~\cite{ejaz2023life}.

\item Anomaly Detection in IoT Networks: With the proliferation of Internet of Things (IoT) devices, AI models monitoring these networks must withstand adversarial attacks that could mask abnormal behavior indicative of a security breach~\cite{wang2303adversarial}.
\end{enumerate}

\subsection{The Reproducibility Crisis in Cybersecurity}

While the importance of adversarial robustness in cybersecurity is clear, the reproducibility crisis poses several risks to the cybersecurity community:

\begin{enumerate}
  \addtolength{\itemindent}{-.3cm}
\item Unreliable Defense Mechanisms: If adversarial robustness techniques cannot be reliably reproduced, it becomes challenging to verify their effectiveness, potentially leading to the deployment of vulnerable AI systems in critical security applications \cite{rudner2021key}.

\item Hindered Progress: The inability to reproduce results impedes the cumulative advancement of knowledge in the field, slowing down the development of more robust AI-driven security solutions.

\item Eroded Trust: As reproducibility issues persist, trust in AI-based cybersecurity solutions may wane, potentially leading to reluctance in adopting these technologies for critical security tasks~\cite{taddeo2019trusting}.

\item Increased Vulnerability: 
As threats rapidly evolve, the inability to quickly validate and build upon existing research could leave organizations vulnerable to new attack vectors.
\end{enumerate}

\section{Methodology}

\subsection{Research Design}

This study aimed to explore challenges in reproducibility of AI-driven cybersecurity research in the context of adversarial robustness. Using a case study methodology, it examined specific issues in the reproduction process, providing insights for broader discussions on reproducibility in cybersecurity.

\citetitle{RN99} \cite{RN99} was chosen for its relevance to AI-driven cybersecurity and the availability of tools and datasets. The goal was to reproduce the results using the 
authors' VeriGauge toolkit~\cite{RN99website}.

\subsection{Attempting to Reproduce the Results from the Previous Paper}

The study followed the original paper's procedures, including software setup and dependency installation. To support VeriGauge's GPU needs, the NVIDIA GeForce RTX 4090 driver on Windows 11 was updated, and Ubuntu 22.04 with the latest NVIDIA CUDA toolkit was installed on WSL2.

The first challenge was that the non-Python dependencies listed in Verigauge/eran/install.sh were outdated or incompatible with current software repositories, necessitating manual sourcing of these packages:

\begin{itemize}
    \item m4-1.4.1.tar.gz
    \item gmp-6.1.2.tar.xz
    \item mpfr-4.2.1.tar.xz
    \item gurobi9.1.0\_linux64.tar.gz
\end{itemize}

The VeriGauge install script otherwise sets most of the necessary environmental variables. Though time-consuming, this process highlights the challenges of rapidly evolving software ecosystems and the importance of preserving specific software versions used in published research.

\subsection{Navigating Software and Hardware Compatibility Issues}

Configuring the experimental environment revealed numerous software and hardware compatibility issues, complicating the reproduction of research across different computational environments. Conflicts between deep learning libraries, CUDA toolkits, and GPU drivers caused cryptic errors and unexpected terminations of the VeriGauge toolkit. Overcoming these issues required exploring alternative hardware, installing specific software versions, and conducting extensive research. This time-consuming process underscores the importance of standardized, well-documented, and containerized environments for consistent execution across diverse platforms.

\begin{table}[H]
  \caption{Python packages sourced in order to run the tests.}
  \label{packages}
  \center
  \resizebox{\columnwidth}{!}{%
\begin{tabular}{|lll|}
\hline
  torch==1.4.0 & torchvision==0.5.0 &numpy==1.18.1  \\
  Keras==2.3.1 & matplotlib==3.1.3 & pycddlib==2.1.1 \\
  pandas==1.0.2 & onnx\_simplifier==0.2.7 & gurobipy==9.1.0 \\
  scipy==1.3.1 & onnxruntime==1.2.0 & cleverhans==3.0.1 \\
  cvxpy==1.1.11 & waitGPU==0.0.3 & numba==0.48.0 \\
  tqdm==4.43.0 & tensorflow-gpu==1.15.4 &  setGPU==0.0.7 \\
  onnx==1.5.0 & setproctitle==1.1.10 & gpustat==0.5.0 \\
  Pillow==7.1.2 & posix\_ipc==1.0.4 & \\
 \hline
\end{tabular}
}
\end{table}

These difficulties are detailed as follows (in more or less chronological order of appearance):

\begin{enumerate}
\item The specified software versions in the Python `requirements.txt' were not all available. For instance, the cdd requirement has a Python 2 style call to print, causing installation to fail under Python 3. To resolve this, the cdd module~\cite{cddwebsite} was downloaded locally, the print statement in setup.py was corrected, and then it was installed into the venv. The finalized versions of requirements that more or less worked are listed in Table \ref{packages}.

\item Versions of TensorFlow newer than TensorFlow 2.0 do not support the contrib module, which appears throughout the codebase. Worse still, TensorFlow 1.x is not available for Python 3.8 or newer. There is a procedure to migrate from TensorFlow 1.x to TensorFlow 2, but it is highly manual.  

\item Related to point 2, Python 3.7 was altinstalled as Python 3.7 and earlier are not available in modern OS repositories.  This version of Python can either be built from scratch or obtained from the Dead Snakes ppa, although both ways require running ``make altinstall'' to avoid disabling the version of Python upon which the operating system depends.  Because this version of Python will conflict with what is already installed on the system, a venv should be used as follows: ``python3.7 -m venv .venv \&\& source .venv/bin/activate''.  This will create a Python 3.7 environment into which the dependencies can be installed.

\item It was discovered that VeriGauge couldn't access the GPU. NVIDIA's website indicated that the installed NVIDIA driver and CUDA toolkit were not backward-compatible with the software. An earlier toolkit version was reinstalled, but the driver wasn't downgraded to maintain compatibility with Windows hosts outside the WSL2 environment.

\item After further investigation, it was discovered that \cite{RN99} conducted all MNIST testing on a GTX 1080 GPU. Therefore, an older system with a GTX 1080 GPU was obtained and re-imaged with Windows 11. WSL2 with Ubuntu 20.04.6 was installed to avoid issues with older software compatibility on Ubuntu 22.04. A backward-compatible driver supporting CUDA 12.2 was installed, and CUDA toolkit version 11.6 was selected as it aligned with the development timeline of VeriGauge.

\item 
Once the system appeared functional, the following CLI (command line interface) command was used to run several tests consistent with \cite{RN99}'s documentation: ``python experiments/evaluate.py --method IBPVer2 --method IBP --method Spectral --method RecurJac --method FastLip --method FastLinSparse --method FastLin''. However, the process soon crashed.  Investigation of /var/log/syslog on the Ubuntu 20.04.6 machine revealed the OOM (out of memory) killer had terminated the process due to RAM exhaustion. Although WSL was reconfigured to access the full 16GB of RAM (from the available 8GB on the host), rerunning the command still resulted in the OOM killer terminating the process.

\item A 64 GB swap space was allocated following \cite{swapwebsite}, and the Python command was rerun.  Processing continued beyond the previous issue, and despite exhausting all physical RAM, the 64GB of swap space was sufficient to continue the computation. However, the Python process eventually attempted to allocate 18.5 GB of GPU RAM on the 8GB RAM causing the OOM killer to terminate the process again.

\item 
The documentation revealed the meaning behind the previous Python command and its components.  It showed that a subset of tests could be run and compared to benchmarks on the related website. A test using the method kReLU on the MNIST dataset using model G, a fully connected neural network with seven layers and 1024 nodes each, was successfully run using the command: ``python experiments/evaluate.py --method kReLU --dataset mnist --model G --mode verify --mode radius --cuda\_ids 0 1$>$/dev/null''. 

\end{enumerate}

During this exhausting journey, reproducing the target study proved elusive due to the compounding effects of software evolution and obsolescence. The research team eventually arrived at a configuration where they could successfully execute a subset of the original results. These experiences serve as a poignant reminder of the urgency with which the reproducibility crisis must be addressed.

\section{Results}\label{results}

\subsection{Successfully Running a Subset of Results}

Despite many obstacles, a subset (16) of the original tests more or less successfully ran. The kReLU method on the MNIST dataset, using a fully connected neural network, was investigated further as it was the first successful test. A successful run produced five log files, listed in Table \ref{results-table}.

\subsection{Inability to Run All Key Tests}

Despite being a relatively recent creation, the VeriGauge toolkit, initially designed for validating and benchmarking adversarial robustness techniques, was itself a victim of the ever-evolving software ecosystem. Outdated or incompatible dependencies made installation and execution difficult. Furthermore, the intricate interplay between the toolkit, deep learning libraries, and hardware dependencies, such as CUDA and GPU drivers, added further challenges. Minor version mismatches or incompatibilities often manifested as cryptic errors or unexpected crashes, hindering reproduction efforts and obscuring the root causes of failures.

\subsection{Discrepancies between Original and Reproduced Results}

Even when reproduction appeared successful, discrepancies between the original and reproduced results emerged. Though seemingly minor, these discrepancies raise concerns about the long-term reproducibility of research due to software evolution. Comparing the original results with the reproduced results required reviewing test log files.

Table \ref{results-table} provides a comparison of the original results obtained by \cite{RN99} and those reproduced by the research team. Column 1 shows the reported results from \cite{RN99} according to the MNIST models - Clean Accuracy section of the benchmark page. Column 2 shows the results from the research team's testing. Column 3 shows the absolute difference between the original reported results and the results of the research team's testing. Column 4 shows the percent difference between the original results and the research team's results.

\begin{table}[H]
  \caption{A Comparison of original versus reproduced results for MNIST - Clean Accuracy.}
  \label{results-table}
  \center
  
  \resizebox{\columnwidth}{!}{%
\begin{tabular}{|lllll|}
  \hline
  Measure & Original \cite{RN99} & Reproduced & Abs. Diff & Rel. Diff.\\
  \hline
  Clean & 95.05\% & 95.00\% & 0.05 & 0.05\%\\
  Adv1 & 98.00\%    & 98.00\% & 0.00 & 0.00\%\\
  Cadv1 & 96.89\% & 96.00\% & 0.89 & 0.92\%\\
  Adv3 & 83.27\% & 78.00\% & 5.27 & 6.32\%\\
  Cadv3 & 35.87\% & 28.00\% & 7.87 & 21.94\%\\
 \hline
\end{tabular}
}
\end{table}

As seen in the table, the ``Clean Accuracy'' results, which represent the baseline performance of the neural network models without adversarial perturbations, exhibited slight deviations from the published values. Some measures exhibited much larger differences, including Adv3 at 6.32\% and Cadv3 at 21.94\%. While some of these deviations are perhaps within an acceptable range for certain applications, some differences are quite profound, and highlight the reproducibility deficiency of the original study. These deviations also suggest the potential for more substantial divergences as software dependencies continue to evolve and the complexity of the verification methods increases. Further, such discrepancies, particularly for Adv3 and Cadv3, suggest potential vulnerabilities in cybersecurity AI models when faced with stronger adversarial attacks.

\section{Discussion}\label{discussion}

\subsection{Challenges}

The challenges encountered in this study underscore the critical nature of the reproducibility crisis in AI-driven cybersecurity research. 
As evident from the narrative in Section \ref{results}, efforts to validate the results from previous work were hindered by numerous technical obstacles, including:

\begin{enumerate}
\item \textit{Software and Hardware Compatibility}: The rapid evolution of software tools and libraries led to version conflicts and obsolescence. Despite the original study being relatively recent, some dependencies specified in the original research were no longer readily available or compatible with modern systems, requiring extensive troubleshooting and configuration adjustments. \cite{mukherjee2021fixing} touches on this issue, describing that frequent updates of Python packages can lead to breaking changes that make historical builds unreproducible.
\item \textit{Documentation Gaps}: Incomplete or outdated documentation exacerbated the reproducibility challenge. Detailed instructions on setting up the experimental environment and running the verification methods were often missing or insufficient, leading to significant time consuming, trial and error. \cite{haibe2020transparency} reports on this issue by pointing to a highly cited paper in which insufficient documentation underlying a study undermines its scientific value by hindering reproducibility.

\item \textit{Resource Intensive Verification}: The computational resources required for complete verification were substantial. The need for high RAM capacity and GPU capabilities, and in a particular configuration, made it difficult to reproduce results on standard hardware, highlighting the need for reproducible, scalable and resource-efficient verification methods.
\end{enumerate}

\section{Proposed Solutions}\label{proposed}

Each of the issues encountered in this study are indicative of the reproducibility challenges faced by the scientific community as a whole. While this paper does not attempt to be the source of solutions for this problem, the following subsections point to related work that aims to address some of the specific challenged faced during this study. 

\subsection{Containerization}

Using container technologies like Docker can standardize computational environments, ensuring consistent code execution across systems and mitigating software version and dependency conflicts. Several research papers highlight the potential of containerization in addressing reproducibility issues in machine learning (ML) and artificial intelligence (AI) by providing shareable, encapsulated environments. For instance, \cite{semmelrock2023reproducibility} discusses the reproducibility crisis in ML research and suggests containerization as a solution, while \cite{boettiger2015docker} emphasizes sharing containers with other researchers. Additionally, \cite{chen2022towards} outlines a systematic approach to training reproducible deep learning models using containers, and \cite{hagmann2023inferential} explores managing and reproducing computational environments for ML experiments.

However, containerization is not a complete solution and should not replace properly documented research. As \cite{raghupathi2022reproducibility} notes, reproducibility ultimately relies on fully documenting methods, data, and experiments.

\subsection{Software Preservation}
   
Establishing repositories for archiving software versions and dependencies used in research is crucial for preserving the computational environment necessary for reproduction. This practice supports long-term reproducibility and allows future researchers to access the original study setups. Several papers advocate for software preservation via repositories to address reproducibility issues. \cite{semmelrock2023reproducibility} emphasizes that sharing code and data alone is insufficient; the entire computational environment must be preserved. \cite{pineau2021improving} discusses challenges and solutions for enhancing reproducibility in ML research, including using repositories like GitHub and the Open Science Framework (OSF) for preserving software and data. \cite{chen2022towards} presents a systematic approach to ensuring the reproducibility of deep learning models by preserving the complete training environment, including code, data, and configuration settings.

\subsection{Comprehensive Documentation}
   
Improving documentation practices is crucial. Researchers should provide detailed, step-by-step instructions for setting up environments, running experiments, and interpreting results, including hardware requirements, software versions, and potential pitfalls. Comprehensive documentation is vital for addressing reproducibility issues in ML and AI research. \cite{albertoni2023reproducibility} offers guidelines emphasizing the need to document code, data, and experimental workflows thoroughly to ensure reproducibility. \cite{pineau2021improving} highlights the importance of documenting methodologies, including code, datasets, and training parameters. \cite{abbas2023rethinking} also stresses the role of thorough documentation in AI research reproducibility.

\subsection{Standardization Efforts}

Standardized frameworks for conducting and reporting research in adversarial robustness can improve reproducibility. This includes creating benchmarks, standardized datasets, and widely accepted evaluation protocols. Several papers address standardization efforts to enhance reproducibility in ML and AI research. \cite{pineau2021improving} highlight the importance of standardized evaluation practices and reporting guidelines. \cite{haibe2020transparency} emphasize that detailed metadata sheets and consistent reporting can help address the reproducibility crisis. \cite{albertoni2023reproducibility} stress the need for standardized documentation, including detailed experimental workflows, data preprocessing, and model training parameters.  

\subsection{Collaborative Initiatives}

Fostering a culture of collaboration and open science is essential for encouraging researchers to openly share their code, data, and methodologies, helping to resolve reproducibility issues. Various sources highlight collaborative initiatives in ML and AI research addressing these challenges. \cite{semmelrock2023reproducibility} emphasizes frameworks and guidelines that promote open collaboration through platforms like GitHub and the Open Science Framework (OSF). \cite{pineau2021improving} discusses community-driven efforts like the ML Reproducibility Challenge, where researchers reproduce and verify results of high-impact ML papers. \cite{abbas2023rethinking} underscores the importance of open-source tools, shared code repositories, and community-driven platforms for validating research.

\section{Momentum toward Addressing Reproducibility}\label{momentum}
The research community must collectively address the reproducibility crisis by prioritizing these best practices and advocating for changes at institutional and policy levels. Funding agencies should support projects that focus on improving reproducibility. Journals and conferences should emphasize the importance of reproducibility in their review processes; artifact review (to help verify reproducibility) and badging (to provide recognition and a stamp of quality) are now commonly a part of that effort. Page length limits, though intended partially to encourage concise writing and ease the burden on reviewers, can inadvertently hinder reproducibility. Researchers often face difficult decisions about which results and discussions to include, leading to the exclusion of valuable details that are crucial for others to replicate their work.

Within the security community, there is a growing dedication to reproducibility in research. For example, in 2023 the ACM SIGSAC Conference on Computer and Communications Security (CCS) introduced an optional artifact evaluation process for all accepted papers \cite{acm2023ccs}. The IEEE Workshop on Offensive Technologies (WOOT) encouraged the submission of artifacts beginning in 2021 \cite{ieee2021woot}. The Usenix Security Symposium began evaluating artifacts of accepted papers in 2020 \cite{usenix2020}.

While these efforts have raised awareness and provided valuable momentum, the AI/ML and security communities have continued unique challenges due to the complexity of the methodologies involved, the rapid evolution of software dependencies, and the critical nature of the applications in which these techniques are employed. Hopefully, with continued attention and effort related to reproducibility, the community can continue to develop tailored solutions and best practices to address the crisis effectively.

\section{Conclusion}\label{conclusion}

The reproducibility crisis in adversarial robustness research, as illuminated by our case study, presents a critical challenge that reverberates throughout the cybersecurity landscape. This crisis not only undermines the scientific integrity of AI research but also poses significant risks to the practical implementation of AI-driven security solutions in an increasingly complex threat environment.

The challenges encountered in our study while attempting to reproduce the results of previous work underscore the urgent need for the cybersecurity research community to adopt best practices that prioritize reproducibility. These practices include containerization to standardize computational environments, comprehensive documentation, and collaborative initiatives to foster transparency and open science. By addressing these issues, researchers can ensure that the AI models designed to protect critical infrastructure and secure networks against threats are more reliably robust.

\section{ACKNOWLEDGMENTS}
ChatGPT assisted with the grammar of this original work.

\printbibliography

\end{multicols}

\end{document}